%% file: main.tex
\documentclass[11pt]{article}

\usepackage{EMNLP2023}

\usepackage{times}
\usepackage{latexsym}

\usepackage[T1]{fontenc}
\usepackage[utf8]{inputenc}

\usepackage{microtype}

\usepackage{inconsolata}

\usepackage{graphicx}
\usepackage{multirow}
\usepackage{tabularx}
\usepackage{subfig}
\usepackage{lineno}
\usepackage{times}
\usepackage[table]{xcolor}
\usepackage{arydshln}
\usepackage{amssymb} %%% changed later
\usepackage{url}
\usepackage{paralist}
\usepackage{makecell}
\usepackage{tikz,lipsum,lmodern}
\usepackage[most]{tcolorbox}
\usepackage{booktabs}
\tcbuselibrary{breakable}
\usepackage{ragged2e}
\usepackage{fontspec}
\usepackage{polyglossia}
\setmainlanguage{english}
\setotherlanguages{hindi,kannada}
\newfontfamily\hindifont[
  Path=fonts/,
  Extension=.ttf,
  UprightFont=*-Regular,
  Script=Devanagari
]{NotoSansDevanagari}
\newfontfamily\kannadafont[
  Path=fonts/,
  Extension=.ttf,
  UprightFont=*-Regular,
  Script=Kannada
]{NotoSansKannada}
\definecolor{fg}{RGB}{34,139,34}

\newcommand{\sv}[1]{\textsc{SamaVaani}}
\newcommand{\iw}[1]{\textsc{IndicWhisper}}
\newcommand{\wl}[1]{\textsc{WhisperLargeV3}}
\newcommand{\sr}[1]{\textsc{Sarvam}}
\newcommand{\gmn}[1]{\textsc{Gemma3n}}
\newcommand{\omni}[1]{\textsc{OmniLingual}}
\newcommand{\va}[1]{\textsc{Vaani}}
\newcommand{\gem}[1]{\textsc{Gemini}}
\newcommand{\ggl}[1]{\textsc{GoogleS2T}}
\newcommand{\stdft}[1]{FT$^\textsc{Std.}$}
\newcommand{\ftps}[1]{FT$^\textsc{PS}$}
\newcommand{\ftcl}[1]{FT$^\textsc{CL}$}
\newcommand{\ftctc}[1]{FT$^\textsc{CTC}$}

\definecolor{lightgreen}{RGB}{235, 247, 237}
\definecolor{lightyellow}{RGB}{255, 249, 230}
\definecolor{langeng}{RGB}{252, 224, 221}
\definecolor{langhin}{RGB}{222, 235, 247}
\definecolor{langkan}{RGB}{229, 245, 224}
\definecolor{gemmaval}{RGB}{233,215,220}
\definecolor{omnival}{RGB}{141,220,220}

\title{\sv{}: Auditing and Debiasing\\ Multilingual Clinical ASR for Indian Languages}

\author{
Subham Kumar\textsuperscript{\dag} Prakrithi Shivaprakash\textsuperscript{\ddag} Abhishek Manoharan\textsuperscript{\ddag} Astut Kurariya\textsuperscript{\ddag} \\ \textbf{Diptadhi Mukherjee\textsuperscript{*}} \textbf{Prabhat Chand\textsuperscript{\ddag}} \textbf{Pratima Murthy\textsuperscript{\ddag}} \\ \textbf{Koustav Rudra\textsuperscript{\dag}} \textbf{Lekhansh Shukla\textsuperscript{\ddag}} \and \textbf{Animesh Mukherjee\textsuperscript{\dag}}  \\
\textsuperscript{\dag}IIT, Kharagpur, \textsuperscript{\ddag}NIMHANS, Bangalore, \textsuperscript{*}LGBRIMH, Tezpur \\
\{kumarshubham209, prakrithishivaprakash, 12.abhishek.m, astutnamo, diptadhimukherjee\}@gmail.com \\ prabhat@vknnimhans.in, \{pratimamurthy, krudra5, drlekhansh, animeshm\}@gmail.com
}

\begin{document}
\maketitle
\begin{abstract}
Automatic Speech Recognition (ASR) is increasingly used to document clinical encounters, yet its reliability in multilingual and demographically diverse Indian healthcare context remains largely unknown. In this study, we first conduct the systematic audit of ASR performance on real-world psychiatric interview data spanning Kannada, Hindi and Indian English, comparing eight state-of-the-art models including \iw{}, \wl{}, \sr{}, \ggl{}, \gmn{}, \omni{}, \va{}, and \gem{}. 
% We evaluate transcription accuracy across languages, speakers, and demographic subgroups, with a particular focus on error patterns affecting patients vs. clinicians and gender-based or intersectional disparities. 
Our results reveal substantial variability across models and languages, with some systems performing competitively in Indian English but failing in regional speech. We further fine-tune two of the best performing open-source models, i.e., \textsc{Gemma3n} and \omni{}, using various methods. With this, we uncover systematic performance gaps tied to speaker role and gender, raising concerns about equitable deployment in clinical settings, which are further mitigated by fairness-aware fine-tuning. To this end, we propose \sv{}, a unified debiasing technique that simultaneously improves ASR performance and improves fairness across demographic groups. %The experimental results on this approach show significant improvement in terms of word error rate and fairness score.}
% By providing a comprehensive multilingual benchmark and fairness analysis, our work highlights the need for culturally and demographically inclusive ASR development for India’s healthcare ecosystem.
\end{abstract}

\input{introduction}

\input{related}
\input{dataset}
\input{method}
\input{experiment}
\input{result}
\input{discussion}
\input{conclusion}

% Entries for the entire Anthology, followed by custom entries
\bibliography{references, reference}
\bibliographystyle{acl_natbib}

\appendix

\input{appendix}

\end{document}

%% file: introduction.tex
\section {Introduction}
The field of psychiatry is highly dependent on language, with a detailed psychiatric interview the primary diagnostic tool \cite{sommers2015clinical} rather than laboratory or radiological investigations.  Subsequently, verbatim transcripts of these interviews are widely used for clinical diagnosis, academic training, qualitative research, and, more recently, for the development of AI systems in psychiatric tasks ~\cite{so2024aligning}. However, producing accurate transcripts remains a major bottleneck: manual transcription is strenuous, time-intensive (5-8 hours for every hour of audio) and prone to human error~\cite{mays2019measuring}. ASR systems provide a scalable alternative albeit transcription errors~\cite{basma2011error} in psychiatric settings can significantly change clinical interpretation~\cite{ciampelli_combining_2023, shikino_clinical_2023}. Modern ASR systems, including proprietary platforms (\ggl{}~\cite{google_speech_to_text}, Microsoft Azure~\cite{azure_speech_to_text}, Amazon Transcribe~\cite{amazon_transcribe}) and open-source models like \textsc{Whisper}~\cite{radford2022whisper} have improved transcription quality. 
% However, these systems do not perform well in conversational, accented, multilingual, and code-mixed speech scenarios common in India~\cite{russell_what_2024}. 
Although these systems work well for standard English (American and British) and in controlled environments, performance deteriorates substantially for non-standard English, conversational and accented English \cite{russell_what_2024}, code-mixing and code-switching, which is common in multilingual contexts like India \cite{sitaram2020surveycodeswitchedspeechlanguage, koenecke2020racial}. In addition, Indian English and Indian regional languages remain underrepresented in global training corpora, leading to lower accuracy and greater variability \cite{javed_svarah_2023,rai_deep_2024}.\\
These difficulties are amplified with clinically distinctive speech patterns in psychiatric interviews. For instance, low tone, hesitations, and long pauses are common in depression~\cite{alpert2001,yang2012,durao2025}; fast and loud speech in mania~\cite{kaczmarekmajer2024,anmella_automated_2024,diflorio2021}; stammering and repeating in anxiety~\cite{silber_appropriate_2018,harrigan1994disfluency,teferra2022}; and disordered agrammatical speech containing neologisms (made-up words to which the patient attaches special meaning) in schizophrenia~\cite{covington2005,stokes2023,hinzen2015}. Beyond this, recordings are often made under acoustically difficult conditions (in wards with noise from ceiling fans, hospital instruments, or ambulance sirens)~\cite{baroudi2026doctor}. Another big concern is the fairness and accuracy of the transcription. These problems may be magnified in psychiatric interviews, as doctors and patients are very different in terms of education, socio-economic background, conversational role, and speaking style~\cite{koenecke2020racial,tatman_gender_2017}. While recent advances in multilingual ASR from open-source models such as \textsc{Whisper}~\cite{radford2022whisper}, \iw{} ~\cite{indicwhisper2024}, \omni{} ~\cite{omnilingualasrteam2025omnilingualasropensourcemultilingual}, \va{}~\cite{vaani_whisper_hi_2025} and commercial models like \sr{}~\cite{sarvam2025} have improved support for multilingual and regional speech in India, existing evaluations rely mainly on general-purpose datasets and do not capture the complexities of real-world psychiatric interviews. Our work addresses this gap by systematically analyzing and further improving the performance of ASR on multilingual psychiatric interactions.
\\
\textbf{Our contributions and findings}: In this work, we present the first systematic audit of ASR systems on real-world multilingual psychiatric interviews in the Indian context. Leveraging a novel dataset spanning Kannada, Hindi, and Indian English, we evaluate eight state-of-the-art ASR models across languages, speaker roles, and demographic groups, and introduce a comprehensive fairness analysis grounded in WER and fine-grained error patterns. We find substantial variability in performance across models and languages, with consistently higher error rates for low-resource languages such as Kannada, as well as systematic disparities across speaker roles and gender. Therefore, we propose \sv{}, a simple yet effective fairness-aware fine-tuning framework combining contrastive learning and CTC alignment, which significantly improves both transcription accuracy (up to $\sim 50\%$ WER reduction) and fairness across demographic groups. Together, our study highlights critical limitations of current ASR systems in clinical settings and provides actionable pathways toward more equitable and robust multilingual ASR deployment in healthcare.

%% file: related.tex
\section{Related work}
\label{sec:related_work}

Research on ASR in Indian multilingual psychiatric settings spans \textit{three} interconnected areas: (i) ASR for psychiatric interviews, (ii) Bias and speaker-level disparities in ASR, and (iii) ASR for Indian languages and accents. We highlight the challenges and gaps in each area that motivate the current study. \\
% \subsection{ASR for psychiatric interviews}
\noindent\textbf{ASR for psychiatric interviews}: 
A growing set of studies has examined the use of ASR to transcribe or analyze psychiatric interviews.~\citet{ciampelli_combining_2023} and~\citet{just_moving_2025}  evaluated ASR in schizophrenia patients, comparing them with healthy controls using interviews conducted in Dutch and German respectively. Several studies~\cite{molina2025automatic,hwang2025evaluating} in Spanish and French psychiatric interviews in patients with psychosis has found high WER performance. Lastly, \citet{seyedi2023using} studied ASR in psychiatric interviews in American English in patients with depression vs. controls with past history of depression, and found no difference in WER in either group. %Collectively, these studies underscore the difficulty of using ASR in real-world psychiatric contexts, but are limited to single and often high-resource languages.
 \\
 \noindent\textbf{Bias and speaker-level disparities in ASR}: 
 Prior works has identified gender differences in YouTube ASR \cite{tatman_gender_2017}, higher error rates for African-American speakers in commercial ASR systems \cite{koenecke2020racial}, and accent-based inequity among non-native German speakers in clinical contexts \cite{just_moving_2025}. Biases related to age, gender, accents and low-resource languages also affect the performance of ASR \citep{feng2024towards}. However, these studies are not designed for multilingual psychiatric interviews, where conversational roles are inherently unequal with clinicians typically producing longer, more structured speech and patients providing shorter and more uncertain responses.%, with psychiatric disorders and medication further affecting speech patterns.}
% ASR disparities across demographic groups are increasingly documented. One study identified gender differences in YouTube ASR, driven by phonetic and prosodic variation \cite{tatman_gender_2017}. \citet{koenecke2020racial} found that major commercial ASR systems produce nearly twice the error rate for African American speakers compared to white speakers. In clinical samples, \citet{just_moving_2025} reported higher error rates for speakers born outside Germany, suggesting accent-based inequities even within a single language group. Age, gender, non-native and regional accents and low-resource languages lead to significant bias in ASR \citep{feng2024towards}. These studies highlight the broader concern that ASR systems are not demographically neutral; however, they do not address multilingual interactions or the socially unequal roles present in psychiatric interviews.
% There remains significant scope for further understanding of how role-based differences influence communication. In healthcare settings, conversation tends to be quite asymmetrical: clinicians typically share longer, well-organized statements, whereas patients often give shorter and sometimes uncertain replies. This is especially true for psychiatric contexts, where disorders and medications themselves impact speech.
% \subsection{ASR for Indian languages and accents}
\\
\noindent\textbf{ASR for Indian languages and accents}: 
Recent work has focused on ASR challenges for Indian English and regional Indian languages. \textit{Svarah} had significantly higher WER for English with Indian accent than for LibriSpeech~\cite{javed2023svarah}.~\citet{rai_deep_2024} observed substantial disparities in gender, region, and speech rate by analyzing 8,740 hours of NPTEL's Indian lecture in English. Considering Indic languages the large datasets \textbf{IndicSUPERB} and \textbf{IndicVoices} further highlight the linguistic, morphological, and prosodic diversity of Indian languages which pose challenges to ASR~\cite{javed_indicsuperb_2022,javed_indicvoices_2024}. However, they do not contain clinical conversations and analysis of error types. In this regard, the Eka Medical ASR Evaluation dataset~\citep{eka_care_eka_2024} offers valuable Indian accents and drug vocabulary with over 3,900 recordings but is limited to brief, static conversations. On the other hand, the \textbf{DISPLACE-M} dataset contains 55 hours of annotated conversational speech in the healthcare domain.  However, this dataset lacks interviews of psychiatric cases. %, which consists of interactions between community healthcare workers and patients - primarily in Hindi, with code-switching to Indian English \citep{meena2026benchmarking}. 

%% file: dataset.tex
\section{Dataset}
\label{sec:dataset}
% \begin{table*}[h!]
%     \caption{Summary of the dataset. ($^*$) indicates median (Q1, Q3) and ($^\#$) indicates Kruskal-Wallis rank sum test.}
%     % \centering\resizebox{.9\textwidth}{!}{
%     \begin{tabular}{cccccc}
%         \toprule
%         \textbf{Characteristics} & \textbf{Overall} & \textbf{English} & \textbf{Hindi} & \textbf{Kannada} & \textbf{p-value$^\#$} \\
%          & N = 162$^*$ & N = 54$^*$ & N = 58$^*$ & N = 50$^*$ \\
%         \midrule
%         Duration & 32.7 & 36.4 & 26.1  & 36.3  & 0.007 \\
%         (minutes) & (12.2, 49.1) & (28.7, 51.1) & (7.6, 44.1) & (4.1, 51.2) &  \\
%         Total words & 4896.5 & 5550.5 & 4226.5 & 4361 & 0.007 \\
%         & (1747, 7293) & (4121, 7677) & (1205, 7470) & (613, 6869) & \\
%         Unique words & 865 & 830 & 730 & 1230 & 0.2 \\
%         & (467, 1259) & (599, 1047) & (405, 1162) & (239, 1678) & \\
%         Moving average & 0.64 & 0.61 & 0.64 & 0.69 & $<$0.001 \\ 
%         type-token ratio & (0.61, 0.67) & (0.59, 0.62) & (0.62, 0.66) & (0.67, 0.71) & \\
%         (window=100) &&&&& \\
%         \bottomrule
%     \end{tabular}
%     % }
%     \label{tab:data_details}
% \end{table*}

\begin{table*}[htbp]
\centering
\scriptsize
\begin{tabular}{lccccc}
\toprule
\textbf{Characteristics} & \textbf{Overall} & \textbf{English} & \textbf{Hindi} & \textbf{Kannada} & \textbf{$p$-value$^\#$} \\
& N = 202$^*$ & N = 54$^*$ & N = 78$^*$ & N = 70$^*$ \\
\midrule
\textbf{Duration} & 30.9 & 36.5 & 25.3 & 27.2 & $<$0.001 \\
 (minutes) & (6.1, 45.6) & (28.7, 51.1) & (4.5, 36.5) & (3.6, 45.2) & \\

\textbf{Total words} & 4756.5 & 5636.5 & 3877.5 & 3637.0 & $<$0.001 \\
 & (1042, 6111.5) & (4331, 6216.8) & (845.8, 7135) & (526, 4873) & \\

\textbf{Unique words} & 1152.0 & 1269.5 & 885.5 & 1450.5 & $<$0.001 \\
 & (407.5, 1412.2) & (1028.8, 1404.2) & (316.2, 1185.5) & (248.8, 1759.8) & \\

\textbf{Moving average} & 0.64 & 0.61 & 0.64 & 0.69 & $<$0.001 \\
\textbf{Type-token ratio} & (0.61, 0.67) & (0.59, 0.62) & (0.62, 0.66) & (0.67, 0.71) & \\
(window = 100) &&&&& \\

\bottomrule
\end{tabular}
\caption{\footnotesize Dataset summary. ($^*$) indicates median (Q1, Q3) and ($^\#$) indicates Kruskal-Wallis rank sum test.}
\label{tab:linguistic_stats}
\end{table*}

The data for this study comes from a tertiary teaching hospital dedicated to the treatment of psychiatric and neurological conditions. The hospital provides free inpatient and outpatient treatment for economically disadvantaged patients, and thus, a majority of beneficiaries are from such a background. We collected 202 audio recordings of patient and doctor/therapist interactions. These recordings were collected using Android mobile phones in mp3 format. While an attempt was made to make recordings in a quiet environment, no special arrangements were made for this. Therefore, the data represent a real-world setting where recordings are made in busy wards and outpatient department rooms. 
The language, duration and lexical diversity of the dataset are summarised in Table ~\ref{tab:linguistic_stats} and the speaker profiles are detailed in Table~\ref{tab:demographics}.
% \begin{table*}[htbp]
% \centering
% \footnotesize
% \caption{Summary of speaker profiles. All demographic categories showed statistically significant differences across languages using the Kruskal-Wallis rank sum test ($p < 0.001$).}
% \label{tab:demographics}
% \begin{tabular}{lcccc}
% \toprule
% \textbf{Characteristics ($\downarrow$)} 
% & \textbf{Overall} 
% & \textbf{English} 
% & \textbf{Hindi} 
% & \textbf{Kannada} \\
% \midrule

% \multicolumn{5}{l}{\textbf{Patient's gender}} \\
% Female (F) 
% & 51 (25.2\%) 
% & 2 (3.7\%) 
% & 18 (23.1\%) 
% & 31 (44.3\%) \\

% Male (M) 
% & 151 (74.8\%) 
% & 52 (96.3\%) 
% & 60 (76.9\%) 
% & 39 (55.7\%) \\

% \midrule

% \multicolumn{5}{l}{\textbf{Doctor's gender}} \\
% Female (F) 
% & 104 (51.5\%) 
% & 54 (100\%) 
% & 30 (38.5\%) 
% & 20 (28.6\%) \\

% Male (M) 
% & 98 (48.5\%) 
% & 0 (0\%) 
% & 48 (61.5\%) 
% & 50 (71.4\%) \\

% \midrule

% \multicolumn{5}{l}{\textbf{Patient's education level}} \\
% $<$Graduate 
% & 132 (65.3\%) 
% & 18 (33.3\%) 
% & 58 (74.4\%) 
% & 56 (80.0\%) \\

% $\geq$Graduate 
% & 70 (34.7\%) 
% & 36 (66.7\%) 
% & 20 (25.6\%) 
% & 14 (20.0\%) \\

% \bottomrule
% \end{tabular}
% \end{table*}

\begin{table*}[htbp]
\centering
\scriptsize
\begin{tabular}{lccccc}
\toprule
\textbf{Characteristics} & \textbf{Overall} & \textbf{English} & \textbf{Hindi} & \textbf{Kannada} & \textbf{$p$-value$^\#$} \\
& N = 202$^*$ & N = 54$^*$ & N = 78$^*$ & N = 70$^*$ \\
\midrule
% \textbf{Patient's age} & 33.0 & 36.0 & 31.0 & 33.5 & 0.87 \\
%  & (27.0, 42.8) & (26.2, 39.0) & (26.0, 45.0) & (29.0, 44.2) & \\
% \textbf{Doctor's age} & 29.0 & 27.0 & 28.0 & 30.0 & $<$0.001 \\
%  & (28.0, 30.0) & (27.0, 27.0) & (28.0, 29.0) & (30.0, 30.0) & \\
% \midrule
\textbf{Patient's gender} & & & & & \\
F & 51 (25.2\%) & 2 (3.7\%) & 18 (23.1\%) & 31 (44.3\%) & $<$0.001 \\
M & 151 (74.8\%) & 52 (96.3\%) & 60 (76.9\%) & 39 (55.7\%) & $<$0.001 \\
\midrule
\textbf{Doctor's gender} & & & & & \\
F & 104 (51.5\%) & 54 (100\%) & 30 (38.5\%) & 20 (28.6\%) & $<$0.001 \\
M & 98 (48.5\%) & 0 (0\%) & 48 (61.5\%) & 50 (71.4\%) & $<$0.001 \\

\midrule
\textbf{Patient's education level} & & & & & \\

$<$Graduate 
& 132 (65.3\%) & 18 (33.3\%) & 58 (74.4\%) & 56 (80.0\%) & $<$0.001 \\

$>=$Graduate 
& 70 (34.7\%) & 36 (66.7\%) & 20 (25.6\%) & 14 (20.0\%) & $<$0.001 \\

\bottomrule
\end{tabular}

\caption{\footnotesize Summary of speaker profiles. ($^*$) Median (Q1, Q3), ($^\#$) Kruskal-Wallis rank sum test.}
\label{tab:demographics}
\end{table*}

This dataset contains speech from 130 unique speakers, including 7 doctors/therapists and 123 patients. All conversations are between two individuals, making 202 unique doctor-patient/therapist-patient pairs.
\\
\textbf{Preprocessing}: All recordings were listened to by two psychiatrists to ensure they were intelligible. It was ensured that the recordings did not contain the name of the patient or any numerical identifier like phone number, etc. However, we did not exclude segments that contained names of places, dates, etc., as we wish to evaluate if such named entities (Section~\ref{sec:discussion}) lead to more errors in ASR.
\\
\textbf{Annotation}: As part of earlier research, transcripts in native languages were available for 103 of these recordings. For the remaining transcripts (99), two psychiatrists transcribed the recordings. The annotation guidelines for transcribing speech into text are given in the Appendix~\ref{apx:annotation_guidelines} with examples for each of the three languages.

%% file: method.tex
\section{Methodological details}
\label{sec:method}
In this section, we first note the base ASR models that we use to perform the audit. We also discuss different fine-tuning approaches to improve the WER and the fairness of the base models. Finally, we introduce the debiasing algorithm used to develop the \sv{} framework.
%In this section, we describe the methodology of the study and the evaluation methods that were employed to assess the findings. Further, we detail the various fine tuning techniques for the best performing open-source model to efficiently align the speech to text phoneme. 
% The procedure for compiling the ASR-generated transcripts is first discussed, followed by a definition of the word error rate (WER) metric and the statistical tests that were performed to ascertain the differences between various ASR models. 
\setlength{\belowdisplayskip}{1pt} \setlength{\belowdisplayshortskip}{1pt}
\setlength{\abovedisplayskip}{1pt} \setlength{\abovedisplayshortskip}{1pt}
% \subsection{Methodology}
% Recall that we have the audio files in an interview format where there are exactly two speakers, i.e., the patient and the doctor. We generate transcripts for each of these audio files with word level timestamps and a standalone transcript as a whole. Four of these  ASR models (OpenAI Whisper, IndicWhisper, Sarvam's Saarika-2.5, and Google's speech-to-text) output word-level timestamps with transcription text, while the remaining four models (Gemma3n, Gemini-2.5-pro, Omnilingual ASR, and Vaani whisper) generate only the transcription text.
\subsection{ASR models}
\label{subsec:asr_models}
\noindent\textbf{Base models}: We evaluate a total of \textit{eight} ASR models, as mentioned earlier. These include \iw{}~\cite{indicwhisper2024}, \wl{}~\cite{openai_whisper_large_v3}, \sr{}~\cite{sarvam2025}, \ggl{}~\cite{google_speech_to_text}, \gmn{}~\cite{gemmateam2025gemma3technicalreport}, \omni{} ~\cite{omnilingualasrteam2025omnilingualasropensourcemultilingual}, \va{}~\cite{vaani_whisper_hi_2025}, and \gem{}~\cite{gemini25pro_2025}. Of these, \ggl{}, \sr{}, and \gem{} are proprietary models inferenced through APIs, while others are open-source.
% The characteristics of these ASR models are listed in Table~\ref{tab:asr_models}.
Recall that we have the audio files in an interview format where each of them has exactly two speakers, i.e., the patient and the doctor. We generate transcripts for each of these long-form audio files. Couple of these ASR models (\sr{}'s Saarika-2.5 and \gem{}) can generate transcripts with long-form audio as input while the other models (\iw{}, \wl{}, \va{}, \gmn{} and \ggl{}) can only transcribe audio in 30-second chunks.

\noindent\textbf{Fine-tuned models}: One of the straightforward ways to improve the overall WER and fairness across demographic groups is fine-tuning the base models. For this we choose two of the best performing open-source ASR models -- \gmn{}\footnote{\url{https://huggingface.co/google/gemma-3n-E4B-it}} and \omni{} that have the best scores across groups (see Section~\ref{sec:results}). We perform two types of fine-tuning as follows.\\
\stdft{}: This refers to the standard LoRA fine-tuning using part of our dataset for training.\\
\ftps{}: Here we double the dataset by augmenting the pitch of the audio files. The hypothesis is that this augmentation of synthetic data would result in stronger fine-tuning, and therefore better WER and fairness. For our experiments, we have used \textsc{PitchShift}\footnote{\url{https://docs.pytorch.org/audio/main/generated/torchaudio.transforms.PitchShift.html}} to augment our original audio by randomly selecting semitones in the range of [$-5, +5$].
\begin{figure}
    \centering
    \includegraphics[width=\linewidth]{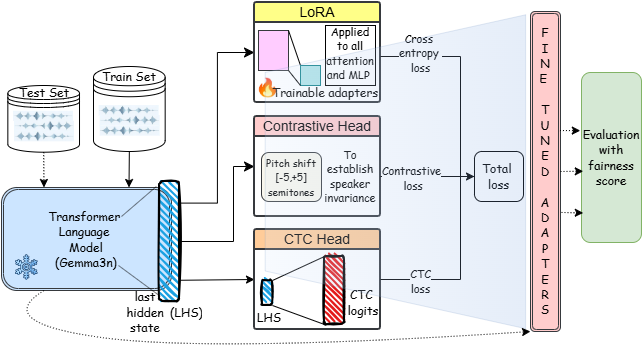}
    \caption{Architecture for proposed fine-tuning pipeline illustrating LoRA, contrastive and CTC head. ($\to$) indicates fine-tuning flow while ($\dashrightarrow$) shows inference on test set.}
    \label{fig:pipeline}
\end{figure}
\subsection{The \sv{} architecture}
While straightforward fine-tuning can potentially improve the WER and the fairness of the models they do not specifically target the issue of debiasing. Recall that the transformer based ASR models follow the modern encoder decoder paradigm where the encoder learns the audio representation and the decoder follows the next word prediction task to generate transcription. Our approach integrates a contrastive learning and a CTC (connectionist temporal classification)~\cite{graves2006ctc} head in the decoding stage while fine-tuning. We only train the LoRA adapters to adapt for Indic languages, freezing the transformer layers as shown in Figure ~\ref{fig:pipeline}. The entire architecture and the associated loss functions are discussed below.\\
\textbf{Contrastive learning}: There are several studies that shows the effectiveness of improving fairness when using a contrastive learning approach ~\cite{koudounas2024contrastive,shen2021contrastivelearningfairrepresentations,9746929_contrastive,ye-etal-2022-contrastive}. As in the case of \ftps{} setup, here also we have used the \textsc{PitchShift} algorithm to augment our original audio by randomly selecting semitones in the range of [$-5, +5$]. Thus, in a batch of $N$ audio samples, we have one original sample and its corresponding pitch-shifted sample constituting a positive pair, and the rest of the $N-1$ audio samples are naturally considered as negative pairs. This forces the model to learn the same utterance regardless of the pitch (a key point of distinction between demographic groups). Further, this doubles the training data and makes the model robust to the high phonetic variance found in languages like Kannada. For an anchor representation $z_a$ and its pitch-shifted positive pair $z_p$, with $N-1$ remaining negative original samples, the contrastive loss for each sample is formulated as:
\begin{equation}
    \mathcal{L}_{CL} = -\log \frac{e^\frac{\text{sim(}z_a, z_p\text{)}}{\tau}}{e^\frac{\text{sim(}z_a, z_p\text{)}}{\tau} \text{ + } \Sigma_{k=1}^{N-1} e^{\frac{\text{sim(}z_a, z_{n,k}\text{)}}{\tau}}}
\end{equation}
where $z_{n,k}$ are the other $N-1$ audio samples in the batch which are considered as negative pairs and $\tau$ is the temperature that dictates the sharpness of the probability distribution, ensuring the model is heavily penalized for any phonetic overlap between the anchor and the negative samples. The contrastive loss implemented is a variation of the NT-Xent (normalized temperature-scaled cross entropy) loss \cite{ntxentloss} which is the standard objective for self-supervised frameworks like SimCLR~\cite{simclr}. Here, we only create the pitch augmentation for the anchor file.\\
\noindent\textbf{CTC head}: In this stage, the logits from last hidden state ($H$) of transformer layers
% \koustav{of what?} 
is projected to the vocabulary space ($V$). By projecting the last hidden states through the CTC head, the model receives a secondary signal that rewards correct character-level sequencing. This head is initialized from scratch and is fully trainable ($W_{CTC}$ = $H\times V$) with CTC loss function represented as, 
\begin{equation}
    \mathcal{L}_{CTC} = -\log P\left(y \mid x\right)
\end{equation}
 where $x$ = ($x_0, x_1, \dots, x_T$) denotes the input sequence of audio features, $T$ is the time steps, $y$ = ($y_0, y_1, ... , y_m$) is the target label sequence text (transcription), and $P\left(y \mid x\right)$ is the total probability of correct transcription over various alignment paths. The CTC loss enables alignment-free training by allowing the model to learn mappings between input frames and output sequences without explicit frame-level labels.\\
\noindent\textbf{Overall loss function}: The weighted sum of the standard cross entropy (CE) loss, the contrastive loss (CL) and the CTC loss constitutes the final loss as follows.
\begin{equation}
    \mathcal{L}_{total} = \mathcal{\alpha \times L}_{CE} \text{ + } \mathcal{\beta \times L}_{CL} \text{ + } \mathcal{\gamma \times L}_{CTC}
    \label{eqn:total_loss}
\end{equation}
where $\alpha, \beta$, and $\gamma$ are the weights for each of the loss components. These weights are optimized using \textsc{Optuna}\footnote{\url{https://optuna.org/}} over 20 trials. The best values of $\alpha, \beta$, and $\gamma$ are selected based on the WER score on the validation set.\\ 
\textbf{LoRA adaptation}: We train all the attention and MLP modules of the transformer based on the above loss function in eq. (~\ref{eqn:total_loss}). This allows the model to repurpose the internal representations for Indic languages with less compute. In addition, it contributes toward the causal cross entropy loss for next word prediction in the speech to text task. Finally, it ensures that the model retains the ability to generate contextually and grammatically correct sentences in text from the speech in a multilingual setting.

%\koustav{Evaluation metric, mplementation details may come under a separate section Experimental Details} \subham{done.}

%% file: experiment.tex
\section{Experimental setup}
\subsection{Evaluation metric}
\noindent\textit{Performance metric}: The main indicator used to assess the accuracy of ASR transcription is the word error rate (WER). Since WER is the most widely used metric for assessing ASRs and has been utilized by several researchers in the literature, we have chosen it as the evaluation metric. The WER is mathematically defined as the ratio of the sum of substitutions ($S$), deletions ($D$), and insertions ($I$) to the total number of words (N) in the reference transcript.
\begin{equation}
\mathcal{WER \%} = \frac{S \text{ + } D \text{ + } I}{N} \times 100
\end{equation}
% It measures the percentage of words incorrectly recognized by the ASR system relative to a ground truth transcript. 
First, we normalize both the ground truth and the generated transcripts by preprocessing the text. This preprocessing includes lowercasing, removing punctuation, and standardizing numbers to reduce superficial mismatches. 
% WER measures ASR performance in terms of word recognition fidelity by taking word-level errors into account. Lower WER indicates better accuracy. 
Second, we use the JiWER Python library for the calculation of $\mathcal{WER}$.\\
\noindent\textit{Fairness metric}: The fairness score ($\mathcal{FS}$) combines a couple of components, namely, the $\mathcal{WER}$ gap and the average $\mathcal{WER}$ among groups (say two groups -- $\mathcal{G}_1$ and $\mathcal{G}_2$). %\koustav{Do we have only two groups?I think it's better to highlight the fairness metric.}
\begin{equation}
    \mathcal{FS} = - \delta \times \mathcal{WER}_{avg} - \theta \times \mathcal{WER}_{gap};  \delta, \theta>=0
\end{equation}
where $\mathcal{WER}_{avg}$ and $\mathcal{WER}_{gap}$ are the average $\mathcal{WER}$ of the two groups ($\mathcal{G}_1$, $\mathcal{G}_2$) and the absolute difference in $\mathcal{WER}$ between the groups ($\mathcal{G}_1$, $\mathcal{G}_2$) respectively. A lower $\mathcal{WER}_{gap}$ suggests better fairness across groups. In addition, a higher value of $\mathcal{FS}$ (ranging from -$\infty$ to 0) indicates overall balanced performance across groups. In this study, we set $\delta = \theta$ = 0.5 to assign equal importance to both units.
\subsection{Implementation details}
We employed \gmn{} and \omni{} multimodal transformer-based models as the backbone for all of our experiments. We implement LoRA by injecting trainable rank 8 decomposition matrices into the query (${q}_{proj}$), key (${k}_{proj}$), value (${v}_{proj}$), output (${o}_{proj}$) and MLP projection layers (${gate}_{proj}$, ${up}_{proj}$, and ${down}_{proj}$). The experimental configuration and the different hyperparameters are noted in Appendix~\ref{apx:hyperparameters}. %Table~\ref{tab:hyperp}. %while maintaining the primary model weights in a 4-bit quantized state to minimize computational overhead. \koustav{This looks clumsy. Can we add a Table at the beginning of the method section to introduce notations?}

%% file: result.tex
\section{Results}
\label{sec:results}
\begin{table}[!ht]
\scriptsize
\centering
% \resizebox{.5\textwidth}{!}{
\scriptsize
\begin{tabular}{ccc}
\toprule
\textbf{Models} & \textbf{WER (\%)} & \textbf{($S, D, I$) \%} \\
\midrule

\multirow{3}{*}{\sr{}}
& \cellcolor{langeng} 34.33 & \cellcolor{langeng} (8.94 ,16.40, 36.87) \\
& \cellcolor{langhin} 39.03 & \cellcolor{langhin} (14.47, 6.76, 39.15) \\
& \cellcolor{langkan} 54.37 & \cellcolor{langkan} (27.87, 17.68, 39.03) \\

\hdashline
\multirow{3}{*}{\ggl{}}
& \cellcolor{langeng} 74.60 & \cellcolor{langeng} (36.13, 40.22, 9.71) \\
& \cellcolor{langhin} 85.55 & \cellcolor{langhin} (10.09, 75.60, 0.13) \\
& \cellcolor{langkan} 94.90 & \cellcolor{langkan} (11.34, 84.01, 0.005) \\

\hdashline
\multirow{3}{*}{\gem{}}
& \cellcolor{langeng} \textbf{14.15} & \cellcolor{langeng} (5.28, 18.58, 4.94) \\
& \cellcolor{langhin} \textbf{18.52} & \cellcolor{langhin} (11.32, 4.35, 8.74) \\
& \cellcolor{langkan} \textbf{35.01} & \cellcolor{langkan} (22.71, 16.86, 7.53) \\

\midrule

\multirow{3}{*}{\wl{}}
& \cellcolor{langeng} 46.76 & \cellcolor{langeng} (9.48, 19.81, 21.76) \\
& \cellcolor{langhin} 71.68 & \cellcolor{langhin} (26.23, 39.17, 9.21) \\
& \cellcolor{langkan} 98.55 & \cellcolor{langkan} (22.22, 76.51, 0.0014) \\

\hdashline
\multirow{3}{*}{\iw{}}
& \cellcolor{langeng} - & \cellcolor{langeng} - \\
& \cellcolor{langhin} 70.3 & \cellcolor{langhin} (23.44, 45.58, 4.27) \\
& \cellcolor{langkan} 97.05 & \cellcolor{langkan} (29.03, 68.21, 0.02) \\

\hdashline
\multirow{3}{*}{\va{}}
& \cellcolor{langeng} - & \cellcolor{langeng} - \\
& \cellcolor{langhin} 44.42 & \cellcolor{langhin} (21.54, 15.60, 6.29) \\
& \cellcolor{langkan} 77.21 & \cellcolor{langkan} (5.56, 24.92, 1.73) \\

\hdashline
\multirow{3}{*}{\gmn{}}
& \cellcolor{langeng} \underline{40.22} & \cellcolor{langeng} (14.63, 17.64, 19.54) \\
& \cellcolor{langhin} 48.14 & \cellcolor{langhin} (19.88, 6.45, 26.48) \\
& \cellcolor{langkan} 90.90 & \cellcolor{langkan} (48.57, 16.67, 30.47) \\

\hdashline
\multirow{3}{*}{\omni{}}
& \cellcolor{langeng} 58.64 & \cellcolor{langeng} (23.40, 31.20, 4.00) \\
& \cellcolor{langhin} \underline{43.55} & \cellcolor{langhin} (23.76, 12.20, 6.53) \\
& \cellcolor{langkan} \underline{75.35} & \cellcolor{langkan} (41.08, 32.19, 2.06) \\

\bottomrule
\end{tabular}
% }
\caption{\footnotesize Overall performance of ASR models across \colorbox{langeng}{English}, \colorbox{langhin}{Hindi}, \colorbox{langkan}{Kannada}. $S$, $D$, $I$ (\%) denote median insertion, deletion, and substitution components of WER. Best results among closed and open-source models are \textbf{bold} and \underline{underlined} respectively.}
\label{tab:overall_perf}
\end{table}
This section is divided into two major parts. In the first part we audit the performance of the eight ASR models across three languages -- Hindi, Kannada and Indian English. Next, we present the performance of the different fine-tuning methods as well as the debiasing algorithm \sv{} proposed by us.
\\
\textbf{Audit outcomes}: Table~\ref{tab:overall_perf} reports the WER and its constituent parts, substitution ($S$), deletion ($D$), and insertion ($I$) in percentages, to compare ASR models in all the three languages. %, namely Indian English, Hindi, and Kannada. 
As we observe from the table, \gem{} achieves the best WER scores (English: 14.15\%, Hindi: 18.52\%, Kannada: 35.01\%), demonstrating better multilingual and generalization abilities.
Among the three languages, Kannada poses relatively higher challenge to the ASR models possibly due to the lack of enough pre-training data. Further, models like \wl{}, \ggl{}, and \gmn{} show a very high WER for Kannada because their architectures and training corpora are not deeply optimised for rich Indian phonetic diversity and retroflex phonemes in Dravidian languages. %Among the open-source models \gmn{} is found to perform the best. 
% We can observe that for the Kannada language, the error mostly stems from deletions (D) reaching as high as 84\% for \ggl{}. The deletions possibly result from the ASR models failing to recognise complex conjugates. Sarvam has an exceptionally high median insertion error (I) in both Hindi (39.15\%) and English (36.87\%). Usually, this indicates hallucinations. Neural ASR models may be caught in a loop (repeating ``and... and... and...'') or produce fluent but inaccurate text based on their internal language model rather than the audio when they are unsure or come across quiet or noise. This increases the number of ``ghost words'', which increases the insertion rate.
\\
\textbf{Fine-tuning and debiasing results}: For fine-tuning, we choose \gmn{} and \omni{} as both have the lowest $\mathcal{WER}$ averaged over three languages. We split the data into train, development (dev) and test folds with language as a stratification variable. Train set had 83.16 hours of audio (English 30.34, Hindi 27.56, Kannada 25.26), dev set had 9.64 hours (English 3.52, Hindi 3.02, Kannada 3.10) and test set had 10.22 hours (English 3.80, Hindi 2.96 and Kannada 3.46 hours).
\setlength{\tabcolsep}{2.5pt}
\begin{table*}[ht]
\centering
\scriptsize
\begin{tabular}{l cc cc cc cc cc cc}
\toprule
\textbf{Metric} 
& \multicolumn{2}{c}{\textsc{Base}} 
& \multicolumn{2}{c}{\stdft{}} 
& \multicolumn{2}{c}{\ftps{}} 
& \multicolumn{2}{c}{\ftcl{}} 
& \multicolumn{2}{c}{\ftctc{}} 
& \multicolumn{2}{c}{\sv{}} \\
\midrule

Overall WER ($\downarrow$)
& \cellcolor{gemmaval}70.47 & \cellcolor{omnival}65.85
& \cellcolor{gemmaval}47.62 & \cellcolor{omnival}47.41
& \cellcolor{gemmaval}41.14 & \cellcolor{omnival}45.83
& \cellcolor{gemmaval}39.08 & \cellcolor{omnival}40.7
& \cellcolor{gemmaval}37.92 & \cellcolor{omnival}38.12
& \cellcolor{gemmaval}\textbf{35.19} & \cellcolor{omnival}\textbf{35.29} \\
\midrule

English WER ($\downarrow$)
& \cellcolor{gemmaval}57.92 & \cellcolor{omnival}47.02
& \cellcolor{gemmaval}25.60 & \cellcolor{omnival}26.02
& \cellcolor{gemmaval}24.13 & \cellcolor{omnival}24.13
& \cellcolor{gemmaval}23.17 & \cellcolor{omnival}24.00
& \cellcolor{gemmaval}22.22 & \cellcolor{omnival}22.44
& \cellcolor{gemmaval}\textbf{20.43} & \cellcolor{omnival}\textbf{20.54} \\

Hindi WER ($\downarrow$)
& \cellcolor{gemmaval}58.33 & \cellcolor{omnival}63.44
& \cellcolor{gemmaval}44.34 & \cellcolor{omnival}43.95
& \cellcolor{gemmaval}38.80 & \cellcolor{omnival}42.31
& \cellcolor{gemmaval}37.25 & \cellcolor{omnival}38.23
& \cellcolor{gemmaval}36.36 & \cellcolor{omnival}36.00
& \cellcolor{gemmaval}\textbf{33.08} & \cellcolor{omnival}\textbf{32.96} \\

Kannada WER ($\downarrow$)
& \cellcolor{gemmaval}84.48 & \cellcolor{omnival}90.36
& \cellcolor{gemmaval}75.00 & \cellcolor{omnival}75.00
& \cellcolor{gemmaval}61.11 & \cellcolor{omnival}72.36
& \cellcolor{gemmaval}58.65 & \cellcolor{omnival}60.94
& \cellcolor{gemmaval}57.82 & \cellcolor{omnival}56.41
& \cellcolor{gemmaval}\textbf{52.84} & \cellcolor{omnival}\textbf{52.88} \\
\midrule

Male (M) ($\downarrow$)
& \cellcolor{gemmaval}69.23 & \cellcolor{omnival}76.55
& \cellcolor{gemmaval}53.33 & \cellcolor{omnival}53.19
& \cellcolor{gemmaval}44.23 & \cellcolor{omnival}50.00
& \cellcolor{gemmaval}41.77 & \cellcolor{omnival}42.31
& \cellcolor{gemmaval}40.71 & \cellcolor{omnival}39.76
& \cellcolor{gemmaval}\textbf{37.25} & \cellcolor{omnival}\textbf{37.25} \\

Female (F) ($\downarrow$)
& \cellcolor{gemmaval}71.43 & \cellcolor{omnival}74.62
& \cellcolor{gemmaval}48.53 & \cellcolor{omnival}48.37
& \cellcolor{gemmaval}42.04 & \cellcolor{omnival}42.86
& \cellcolor{gemmaval}39.41 & \cellcolor{omnival}42.31
& \cellcolor{gemmaval}38.33 & \cellcolor{omnival}39.08
& \cellcolor{gemmaval}\textbf{34.83} & \cellcolor{omnival}\textbf{35.48} \\
\midrule

$\mathcal{FS}$: M vs F ($\uparrow$)
& \cellcolor{gemmaval}-36.27 & \cellcolor{omnival}-43.76
& \cellcolor{gemmaval}-27.87 & \cellcolor{omnival}-27.80
& \cellcolor{gemmaval}-22.66 & \cellcolor{omnival}-26.79
& \cellcolor{gemmaval}-21.48 & \cellcolor{omnival}-22.46
& \cellcolor{gemmaval}-20.95 & \cellcolor{omnival}-20.05
& \cellcolor{gemmaval}\textbf{-19.23} & \cellcolor{omnival}\textbf{-19.07} \\
\midrule

Patient (P) ($\downarrow$)
& \cellcolor{gemmaval}76.47 & \cellcolor{omnival}84.44
& \cellcolor{gemmaval}53.85 & \cellcolor{omnival}53.85
& \cellcolor{gemmaval}47.62 & \cellcolor{omnival}49.04
& \cellcolor{gemmaval}45.45 & \cellcolor{omnival}47.49
& \cellcolor{gemmaval}44.71 & \cellcolor{omnival}43.40
& \cellcolor{gemmaval}\textbf{41.67} & \cellcolor{omnival}\textbf{41.18} \\

Doctor (D) ($\downarrow$)
& \cellcolor{gemmaval}63.16 & \cellcolor{omnival}66.92
& \cellcolor{gemmaval}50.00 & \cellcolor{omnival}50.00
& \cellcolor{gemmaval}41.54 & \cellcolor{omnival}46.18
& \cellcolor{gemmaval}39.24 & \cellcolor{omnival}41.46
& \cellcolor{gemmaval}37.60 & \cellcolor{omnival}38.03
& \cellcolor{gemmaval}\textbf{34.40} & \cellcolor{omnival}\textbf{34.69} \\
\midrule

$\mathcal{FS}$: P vs D ($\uparrow$)
& \cellcolor{gemmaval}-41.56 & \cellcolor{omnival}-51.60
& \cellcolor{gemmaval}-27.88 & \cellcolor{omnival}-27.88
& \cellcolor{gemmaval}-25.33 & \cellcolor{omnival}-26.48
& \cellcolor{gemmaval}-24.28 & \cellcolor{omnival}-25.25
& \cellcolor{gemmaval}-24.13 & \cellcolor{omnival}-23.04
& \cellcolor{gemmaval}\textbf{-22.65} & \cellcolor{omnival}\textbf{-22.21} \\
\midrule

$>$=Graduate ($\downarrow$)
& \cellcolor{gemmaval}63.64 & \cellcolor{omnival}75.23
& \cellcolor{gemmaval}28.45 & \cellcolor{omnival}28.28
& \cellcolor{gemmaval}26.66 & \cellcolor{omnival}28.45
& \cellcolor{gemmaval}25.16 & \cellcolor{omnival}27.27
& \cellcolor{gemmaval}25.16 & \cellcolor{omnival}26.43
& \cellcolor{gemmaval}\textbf{22.28} & \cellcolor{omnival}\textbf{22.42} \\

$<$Graduate ($\downarrow$)
& \cellcolor{gemmaval}80.00 & \cellcolor{omnival}83.44
& \cellcolor{gemmaval}70.43 & \cellcolor{omnival}70.47
& \cellcolor{gemmaval}61.53 & \cellcolor{omnival}64.80
& \cellcolor{gemmaval}59.17 & \cellcolor{omnival}62.17
& \cellcolor{gemmaval}58.82 & \cellcolor{omnival}57.14
& \cellcolor{gemmaval}\textbf{53.64} & \cellcolor{omnival}\textbf{53.39} \\
\midrule

$\mathcal{FS}$: $>$=G vs $<$G ($\uparrow$)
& \cellcolor{gemmaval}-44.09 & \cellcolor{omnival}-48.07
& \cellcolor{gemmaval}-45.71 & \cellcolor{omnival}-39.81
& \cellcolor{gemmaval}-39.49 & \cellcolor{omnival}-38.90
& \cellcolor{gemmaval}-38.09 & \cellcolor{omnival}-36.81
& \cellcolor{gemmaval}-37.83 & \cellcolor{omnival}-36.25
& \cellcolor{gemmaval}\textbf{-34.66} & \cellcolor{omnival}\textbf{-34.44} \\

\bottomrule
\end{tabular}
\caption{\footnotesize Comparison of WER scores of \colorbox{gemmaval}{\gmn{}} and \colorbox{omnival}{\omni{}} across different fine-tuning techniques and different demographic attributes. Lower WER ($\downarrow$) and higher $\mathcal{FS}$ ($\uparrow$) indicates better performance. \ftcl{}: An ablation of \sv{} where only the contrastive loss is used. \ftctc{}: An ablation of \sv{} where only the CTC loss is used. \textbf{Bold} indicates the best score obtained from one model.}
\label{tab:fair_wer}
\end{table*}

\noindent\textit{Main results}: We present the results from the different fine-tuned models in Table~\ref{tab:fair_wer}. The results are organized under different categories including overall $\mathcal{WER}$, language-wise $\mathcal{WER}$ and various demographic-wise $\mathcal{WER}$. Further we also present the fairness score ($\mathcal{FS}$) for each demographic group. The key observations are as follows.
\begin{compactenum}
    \item We observe that \sv{} results in a reduction of 50\% in overall $\mathcal{WER}$ when compared to the \textsc{Base} pre-trained model. The overall $\mathcal{WER}$ of \sv{} is also substantially better than standard fine-tuning setups \stdft{} and \ftps{}.
    \item Across all the three languages \sv{} is remarkably better than the \textsc{Base} model as well as \stdft{} and \ftps{}. Among the three languages, even \sv{} struggles the most with Kannada, like all the other models.
    \item For all the demographic groups \sv{} is not only better in terms of the group-wise $\mathcal{WER}$ but also in terms of $\mathcal{FS}$ when compared to \textsc{Base}, \stdft{} and \ftps{}.
\end{compactenum}
\noindent\textit{Ablation study}: A natural question in the design of \sv{} regards the necessity of both the CL and CTC heads. In order to check whether any one of them is as good as the combination, we present in columns 4 and 5 of Table~\ref{tab:fair_wer} the results from two additional fine-tuning setups -- (i) \ftcl{}: an ablation of \sv{} where only the contrastive loss is used and (ii) \ftctc{}: an ablation of \sv{} where only the CTC loss is used. We observe that though these models outperform the standard fine-tuning setups \stdft{} and \ftps{} in terms of both $\mathcal{WER}$ and $\mathcal{FS}$, they are not as good as \sv{}. This quantitatively justifies the benefit of combining the two loss terms. In the next section, we present some qualitative insights into the advantages of each of these components.  
%We investigate whether contrastive learning (CL) and CTC head independently helps improve the WER and fairness of the system in the given setup of conversational data. The experiment examines the contribution of contrastive learning and CTC head. The columns 4 and 5 shows how the performance of \textsc{Gemma3n} varies when fine-tuned only with CL (\textsc{Gemma3n-FT-cl}) and CTC (\textsc{Gemma3n-FT-ctc}) independently. The results in columns 4-6 of Table~\ref{tab:fair_wer} demonstrate that the CTC head provides essential character alignment from speech to text. In addition, the contrastive learning objective acts as a phonetic regularizer. The combination of these two techniques allows the model to overcome the limitations of autoregressive fine-tuning, essentially for complex multilingual and multi-demographic ASR tasks.

%% file: discussion.tex
\section{Discussion}
\label{sec:discussion}
\begin{table*}[h]
\centering
\scriptsize
\begin{tabular}{p{3cm} p{13cm}}
\toprule
\textbf{Type} & \textbf{Transcript} \\
\midrule

\textbf{Ground truth} & friends, relatives. Yeah, if he values $\dots$ if he values that conflicted person a very high level in in his own mind $\dots$ it is uh, good decision that to worry $\dots$ for him $\dots$ that he didn't come. If he is a normal person who had a conflict, he won't mind that. He won't mind that. \\
\midrule
\textsc{Base} & friends, \textcolor{red}{celebrities}. Yeah, if \textcolor{red}{if if if if $\dots$ (repeated 210 times more)}\\

\stdft{} & friends, \textcolor{red}{celebrities}. Yeah, if \textcolor{red}{if if if if $\dots$ (repeated 212 times more)} \\

\ftcl{} & friends, \textcolor{red}{celebrities}. Yeah, if \textcolor{red}{if if if if $\dots$ (repeated 212 times more)} \\
\midrule
\ftctc{} & friends, \textcolor{red}{celebrities}. Yeah, if \textcolor{red}{if if if you value} if \textcolor{red}{if you value} that conflicted person a very high level in in his own mind, it is a good decision that to worry for him that he didn't come. If he is a normal person who had a conflict, he won't mind that. He won't mind that. \\
\midrule
\sv{} & friends \textcolor{red}{cigarettes}. Yeah, if you value if \textcolor{red}{if you value} that conflicted person a very high level in in his own mind, it is \textcolor{red}{a} good decision that to worry for him that he didn't come. If he is a normal person who had a conflict, he won't mind that. He won't mind that. \\

\bottomrule
\end{tabular}
\caption{\footnotesize Qualitative comparison of transcriptions across models of \gmn{}. Here the speaker is a male patient.}
\label{tab:repeating_words}
\end{table*}

In this section, we discuss representative qualitative advantages of the CL and CTC loss terms (for \gmn{}\footnote{Similar observations hold for \omni{}.}) and finally discuss some error cases. We posit that while the CTC head provides essential character alignment from speech to text, the contrastive learning objective acts as a phonetic regularizer. \\
\textit{Role of contrastive learning}: Recall that we use \textsc{PitchShift} to obtain a pitch-shifted variant of the anchor audio. With this, the objective of contrastive learning is to recognize the semantic equivalence between the original audio and its pitch augmented variant to ignore acoustic noise and focus on phonetic content. By restricting the augmentation to a single anchor-positive pair ($z_i, z_i^+$) within a pool of $N-1$ negative samples ($z_k$), the framework creates an asymmetric learning signal that is highly effective for phonetically dense languages. We set the temperature ($\tau$) at 0.05 to sharpen the distribution, forcing the model to be very certain about the representations of samples in the latent space. $\mathcal{FS}$ improve consistently on all demographic attributes by 13-41\% compared to the base model and 15-22\% compared to the standard fine-tuned model.\\
\textit{Role of CTC head}: Standard LLMs use auto regressive decoding, which are prone to hallucinations with same words repeating sequentially, for example, \textcolor{fg}{if} \textcolor{red}{ if if if...} (get caught in a word repeating indefinitely). The CTC heads enforces monotonic alignment. %The CTC head consists of a fully connected linear layer projected from the last hidden state to the vocabulary space of the model to have character-level temporal alignment. It ensures that the predicted characters or tokens follow in the correct order as the audio progresses in time. For training, the vocabulary logits are passed directly into the \textsc{nn.CTCLoss} function. The loss is calculated by comparing these logits against the ground-truth transcriptions. However at inference, the argmax (or a beam search) on the vocab size determines which character or token is most likely.
While the generative head ensures the sentence makes sense, the CTC head acts as a ``sanity check'' to ensure every word/token corresponds to an actual speech in the file. This balance significantly reduces instances of ``word-skipping'' or adding ``filler'' that wasn't in the original audio.
In the raw audio signal, a single phoneme (like `s' in ``speech'') lasts for many frames. Without a special mechanism, a model might predict the letter `s' many times in a row. CTC solves this using a unique decoding rule involving blank tokens ($\phi$).
The CTC decoding algorithm follows two simple but powerful rules to convert a long sequence of frame-by-frame predictions into a clean word as follows -- (i) \textit{collapse identical consecutive tokens}: if the model predicts `aaaa-bbbb-cccc' due to slow speech, CTC collapses them into `abc'; (ii) \textit{blank as a separator}: to actually output two of the same letter (like the `ll' in `hello'), the model must predict a blank token between them (e.g., h-e-l-$\phi$ -l-o).\\
%CTC is limited by the length of the audio. It cannot predict more tokens than there are frames in the hidden states unlike any standard auto regressive model which can generate text foreover. Furthermore, when there is silence or noisy data, CTC is trained to predict the blank token. This effectively mutes the output during non-speech segments, preventing it from hallucinating filler text during pauses.
\textit{Error analysis}: Table~\ref{tab:repeating_words} shows qualitative examples as to how auto-regressive models can get stuck in a loop where they start repeating the same word sequentially. Fine-tuning with only contrastive learning also fails to get out of the repeating loop. On the other hand, incorporating a CTC head on top LoRA is able to break out of this loop and generates better text. Furthermore, \sv{} generates the best transcripts compared to ground truth. Not only it can get out of the indefinite loop, it also removes the same repeating words with a few inconsistencies. Lastly, the error in the transcripts generated by \sv{} is essentially divided into the following three categories. 
\begin{compactenum}
    \item \textit{Inverse text normalization}: The generation should appear in text as spoken word-by-word. For instance, ``So, no matter what time I sleep, \textcolor{red}{8:00} o'clock is when I have to wake up.'' Here, the time should appear in words \textcolor{fg}{`eight'} as it is spoken and not as it is represented.
    \item \textit{Named entity errors}: There are several instances where the model fails to correct text for named entities, especially for organization and drug names. For example, the organization name \textcolor{fg}{`NIMHANS'} in ground truth is being substituted by terms like \textcolor{red}{`neeman's'} or \textcolor{red}{`2 months'}. Similarily, the drug name  \textcolor{fg}{`benzodiazipines'} is being transcribed as \textcolor{red}{`benzodiazepam pains'}, suggesting a lack of acoustic robustness, where the model struggles to align the correct token sequences.
    \item \textit{Deletion errors}: There are a few missing phonemes while transcribing speech to text. Forcing monotonic alignment with the CTC head prevents the model from skipping any word/token. It forces the model to attempt a phonetic transcription of every sound.
\end{compactenum}

%% file: conclusion.tex
\section{Conclusion}
\label{sec:conclusion}

In this study, we highlighted the disparities in ASR models, particularly in a clinical psychiatric interview setting between a patient and a doctor. We comprehensively audited eight state-of-the-art models across three linguistically diverse languages and reported multiple disparities. Next, we propose \sv{} that leads to simultaneous improvement of the overall $\mathcal{WER}$ as well as the demographic fairness. The key uniqueness lies in combining the two loss functions based on contrastive learning and CTC that serve complementary roles.

While our approach is better than the other fine-tuning setups, there is still a lot of scope for improvement, especially for languages like Kannada. The values suggest that it is important to build customised ASR models for a clinical psychiatric interview setting, pre-trained from scratch. 

%Our findings suggest that almost all models have WER scores lowest for English, highest for Kannada, with Hindi in the middle (except for \omni{}). However, this may occur due to the bias in their training data. The WER scores for each language across models is considered moderate to high which suggests the need to develop an ASR model for clinical psychiatric interview setting, trained from scratch.

\section{Limitations}
While this study poses a solid foundation towards multilingual ASR in the psychiatric interview setting, there are a couple of limitations. \textbf{\textit{First}}, the experimental setup depends on LoRA fine-tuning with rank of 8 (\textit{low}). This setup was chosen due to our hardware constraints. With a high-end infrastructure, a full supervised fine-tuning may further improve the transcription accuracy and fairness. And \textbf{\textit{second}}, our evaluation is limited only to Indian English, Hindi and Kannada psychiatric interviews. India contains substantial linguistic diversity, and ASR behaviour may differ considerably across other Indic languages, dialects, and code-mixed settings.

\section{Ethical consideration}
This study was approved by the Institute Ethics Committee. The speech data were collected from a tertiary teaching hospital with a specialised addiction treatment centre offering 24-hour emergency services dedicated to the treatment of psychiatric and neurological conditions, along with inpatient and outpatient services. Written informed consent was obtained from patients to audio-record psychiatric interviews. Although deidentified, the data cannot be made public as it contains highly sensitive personal health information, which can compromise patient privacy and confidentiality.

% The patients agreed on the written consent to record their case scenarios in a formal clinical setting. We do not remove or filter any personal identifiers from the speech data. As a result, the data cannot be made public as this may compromise patient privacy and confidentiality.

% \section{Funding support}
% This pilot project is supported by an award from Neuromatch, Inc. as part of the Generative AI for Mental Health Research Accelerator, funded by Wellcome Trust Limited. [Grant number: MEXA2025-003].

%% file: appendix.tex
\section{Implementation details \& hyperparameters}
\label{apx:hyperparameters}
We implement LoRA by injecting trainable rank 8 decomposition matrices into the query (${q}_{proj}$), key (${k}_{proj}$), value (${v}_{proj}$), output (${o}_{proj}$) and MLP projection layers (${gate}_{proj}$, ${up}_{proj}$, and ${down}_{proj}$) on the base ASR models. The experimental configuration and the different hyperparameters are noted in Table~\ref{tab:hyperp}. %while maintaining the primary model weights in a 4-bit quantized state to minimize computational overhead. \koustav{This looks clumsy. Can we add a Table at the beginning of the method section to introduce notations?}
%The scaling factor of 16 (usually double the rank to balance the training stability) is set to provide sufficient capacity to capture complex phonetic structures in Kannada and Hindi. A dropout rate of \textbf{0.09} is applied to the adapter layers to mitigate from overfitting. In addition, the \textbf{AdamW} 8-bit optimizer with a base learning rate of $5 \times 10^{-5}$ is used for optimization, following a cosine learning rate schedule with a \textbf{10\%} warmup period for \textbf{3} epochs. We set a maximum sequence length of \textbf{1024} tokens to handle the diverse lengths of the audio-transcription pairs. The coefficient weights in the total loss (eq.~\ref{toal_loss}) are optimized using \textsc{Optuna} over \textbf{20} trials. The values of $\alpha, \beta$, and $\gamma$ are obtained as \textbf{0.4135}, \textbf{0.2186}, and \textbf{0.3679} respectively. The temperature parameter ($\tau$) in contrastive learning objective is set to \textbf{0.05}, forcing the model to distinguish between subtle phonetic variations. Fine tuning is done with a per-device batch size of \textbf{1} and a gradient accumulation factor of \textbf{8}, resulting in an effective global batch size of \textbf{32} across \textbf{4} NVIDIA A6000 GPUs cores of \textbf{48GB} each. Model performance is monitored via Median WER on a validation set, with an early stopping patience of \textbf{3} evaluation steps to ensure robust convergence.

\begin{table}[ht]
\centering
\footnotesize
\renewcommand{\arraystretch}{1.3}
\begin{tabular}{ll}
\toprule
\textbf{Configuration} & \textbf{Value} \\
\midrule
\multicolumn{2}{l}{\textit{LoRA configuration}} \\
\quad Rank ($r$) & 8 \\
\quad Scaling factor ($\alpha_{\text{LoRA}}$) & 16 \\
\quad Dropout & 0.09 \\
\quad Target modules & \makecell[l]{$q_{\text{proj}}, k_{\text{proj}}, v_{\text{proj}}$,\\ $o_{\text{proj}}, gate_{\text{proj}}$,\\ $up_{\text{proj}}, down_{\text{proj}}$} \\
\quad Base model quantization & 4-bit \\
\midrule
\multicolumn{2}{l}{\textit{Optimization}} \\
\quad Optimizer & AdamW (8-bit) \\
\quad Base learning rate & $5 \times 10^{-5}$ \\
\quad Learning rate schedule & Cosine \\
\quad Learning rate warmup period & 10\% (for 3 epochs)\\
%\quad Epochs & 3 \\
\quad Max sequence length & 1024 tokens \\
\quad Temperature ($\tau$) & 0.05 \\
% \midrule
% \multicolumn{2}{l}{\textit{Loss coefficients (tuned via \textsc{Optuna}, 20 trials)}} \\
% \quad $\alpha$ & 0.4135 \\
% \quad $\beta$ & 0.2186 \\
% \quad $\gamma$ & 0.3679 \\
% \quad Temperature ($\tau$) & 0.05 \\
\midrule
\multicolumn{2}{l}{\textit{Training infrastructure}} \\
\quad Per-device batch size & 1 \\
\quad Gradient accumulation steps & 8 \\
\quad Effective global batch size & 32 \\
\quad GPUs & \makecell[l]{4 $\times$ NVIDIA A6000\\ (48\,GB each)} \\
\midrule
\multicolumn{2}{l}{\textit{Validation \& early stopping}} \\
\quad Validation metric & Median WER \\
\quad Early stopping patience & 3 evaluation steps \\
\bottomrule
\end{tabular}
\caption{\label{tab:hyperp}Common experimental configuration and hyperparameters.}
\label{tab:hyperparams}
\end{table}

The loss coefficients ($\alpha, \beta$, $\gamma$) for each of the three components in the total loss function for both the fine-tuned models, \gmn{} and \omni{} are optimized through \textsc{Optuna}\footnote{\url{https://optuna.org/}} over 20 trials. The exact coefficient values for \gmn{} and \omni{} are (0.4135, 0.2186, 0.3679) and (0.4385, 0.2480, 0.3135) respectively.

\section{Acoustic features and fairness analysis of dataset}
\begin{figure}[t]
  \includegraphics[width=\linewidth]{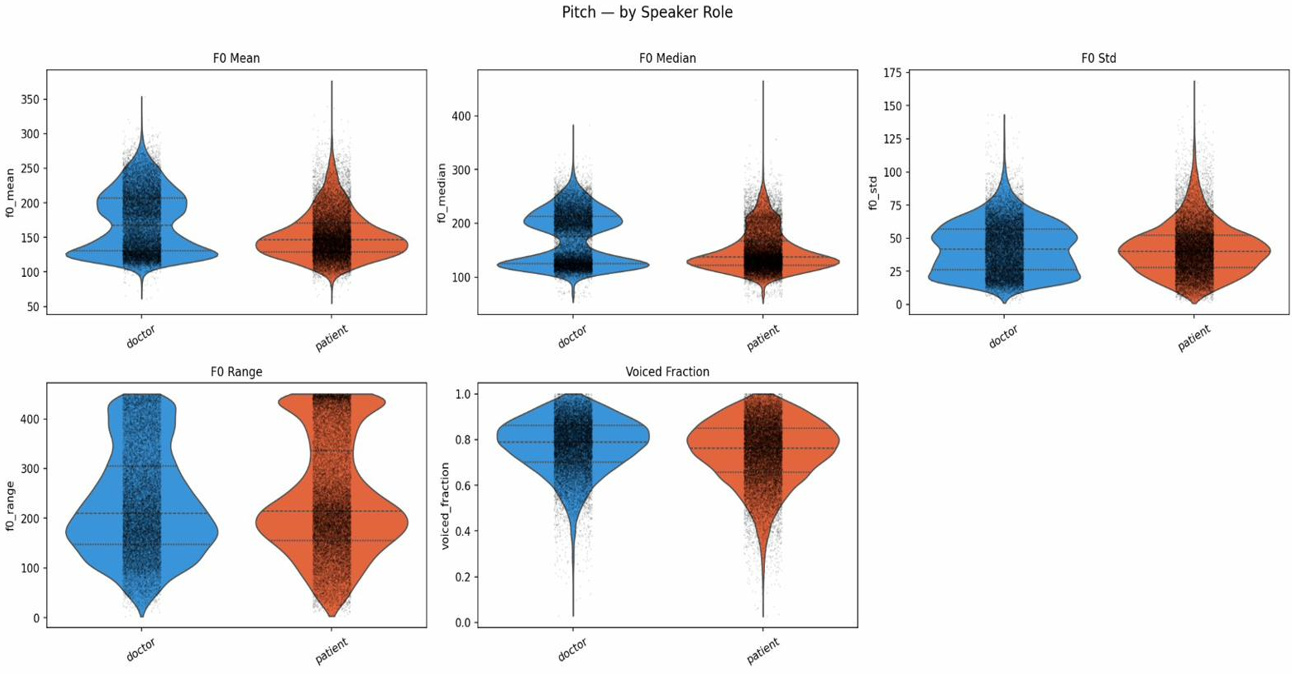}
  \caption{\footnotesize Pitch comparison based on the role of the speaker, i.e, doctor or patient.}
  \label{fig:pitchbyrole}
\end{figure}

\begin{figure}[t]
  \includegraphics[width=\linewidth]{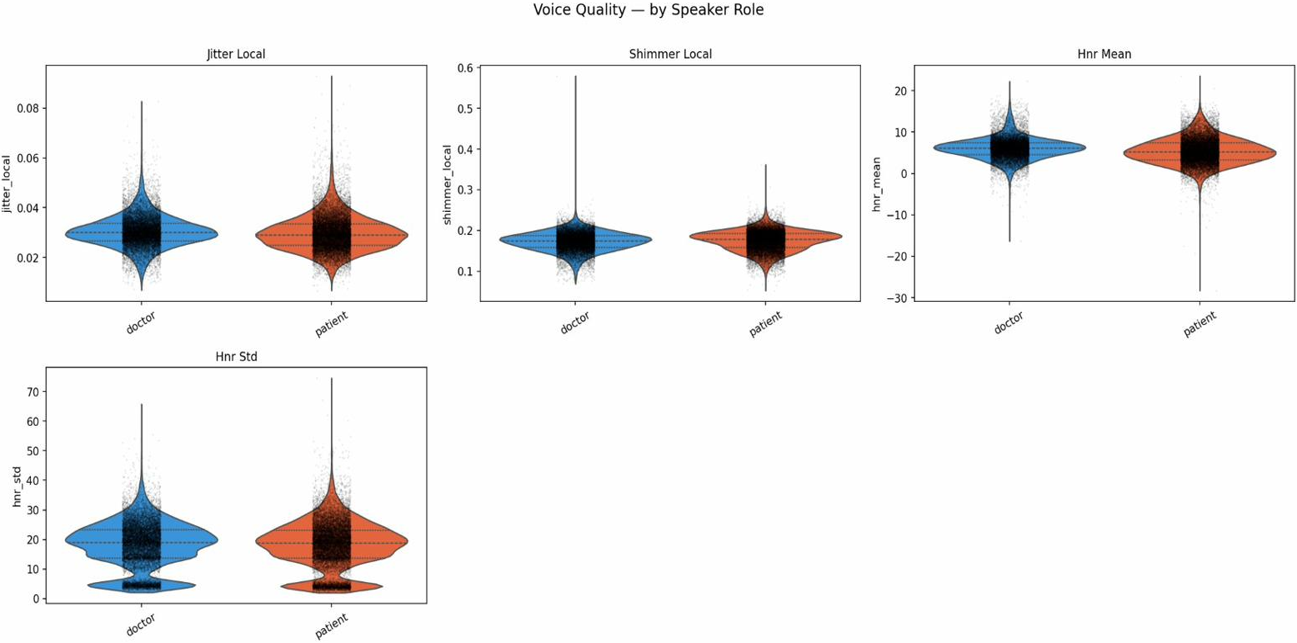}
  \caption{\footnotesize Voice quality comparison based on the role of the speaker, i.e, doctor or patient.}
  \label{fig:vcbyrole}
\end{figure}

\begin{figure}[t]
  \includegraphics[width=\linewidth]{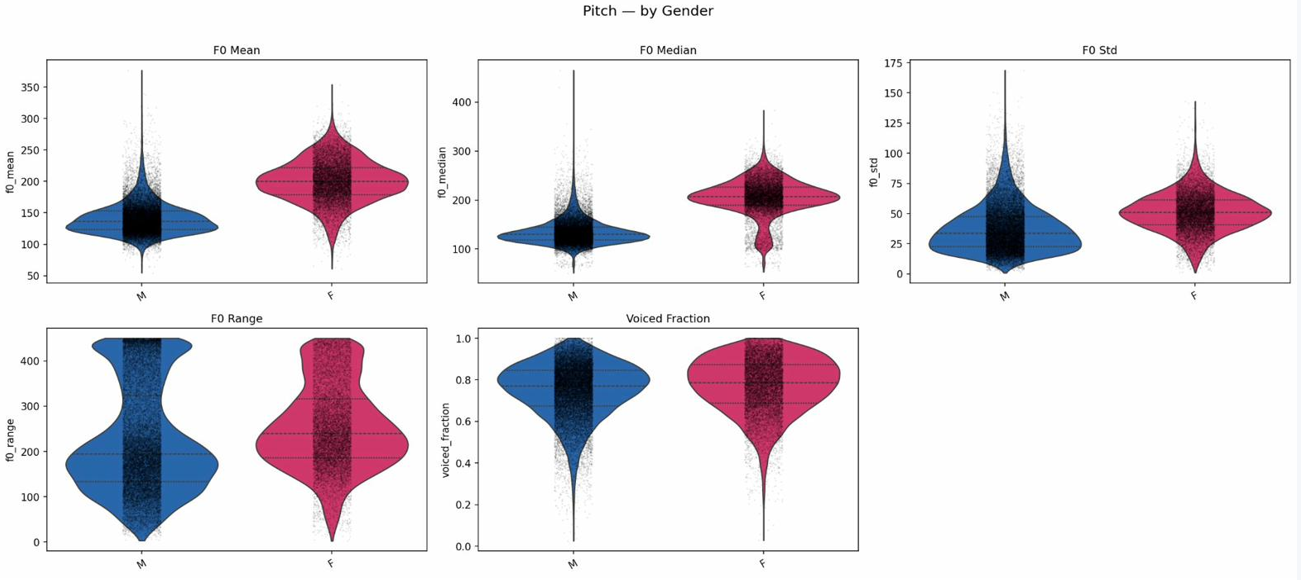}
  \caption{\footnotesize Pitch comparison based on the gender of the speaker, i.e, male or female.}
  \label{fig:pitchbygender}
\end{figure}

\begin{figure}[t]
  \includegraphics[width=\linewidth]{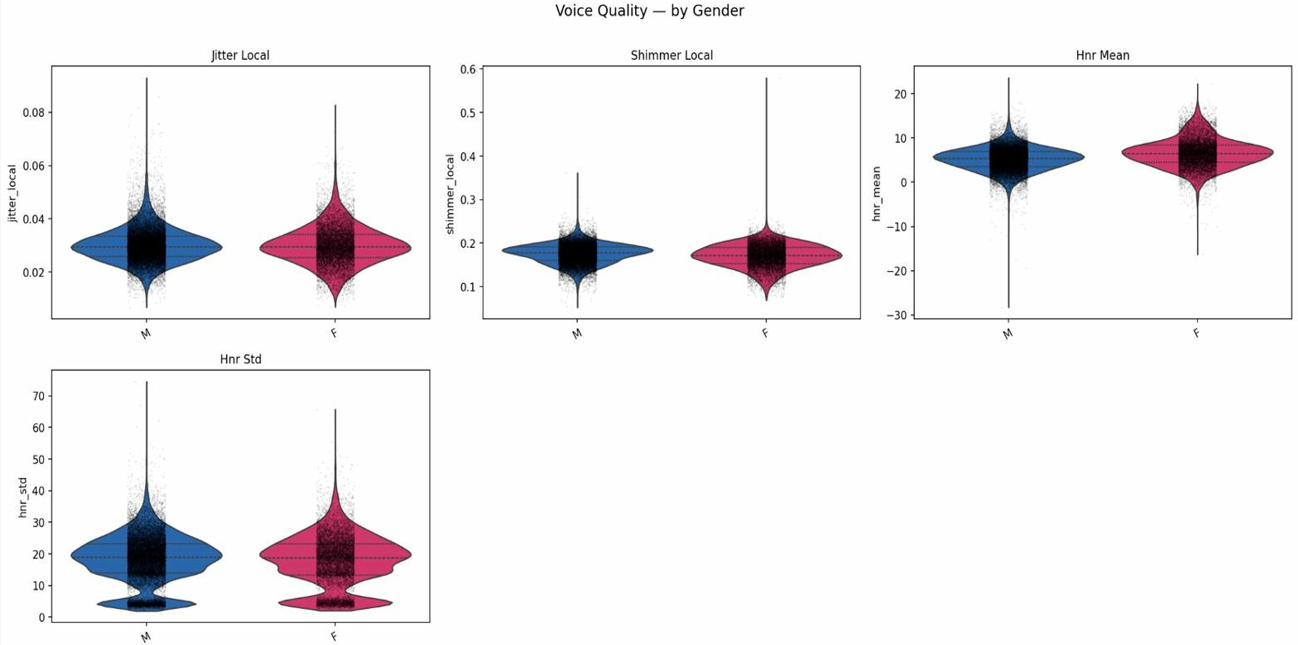}
  \caption{\footnotesize Voice quality comparison based on the gender of the speaker, i.e, male or female.}
  \label{fig:vcbygender}
\end{figure}

\begin{figure*}[t]
  \includegraphics[width=\linewidth]{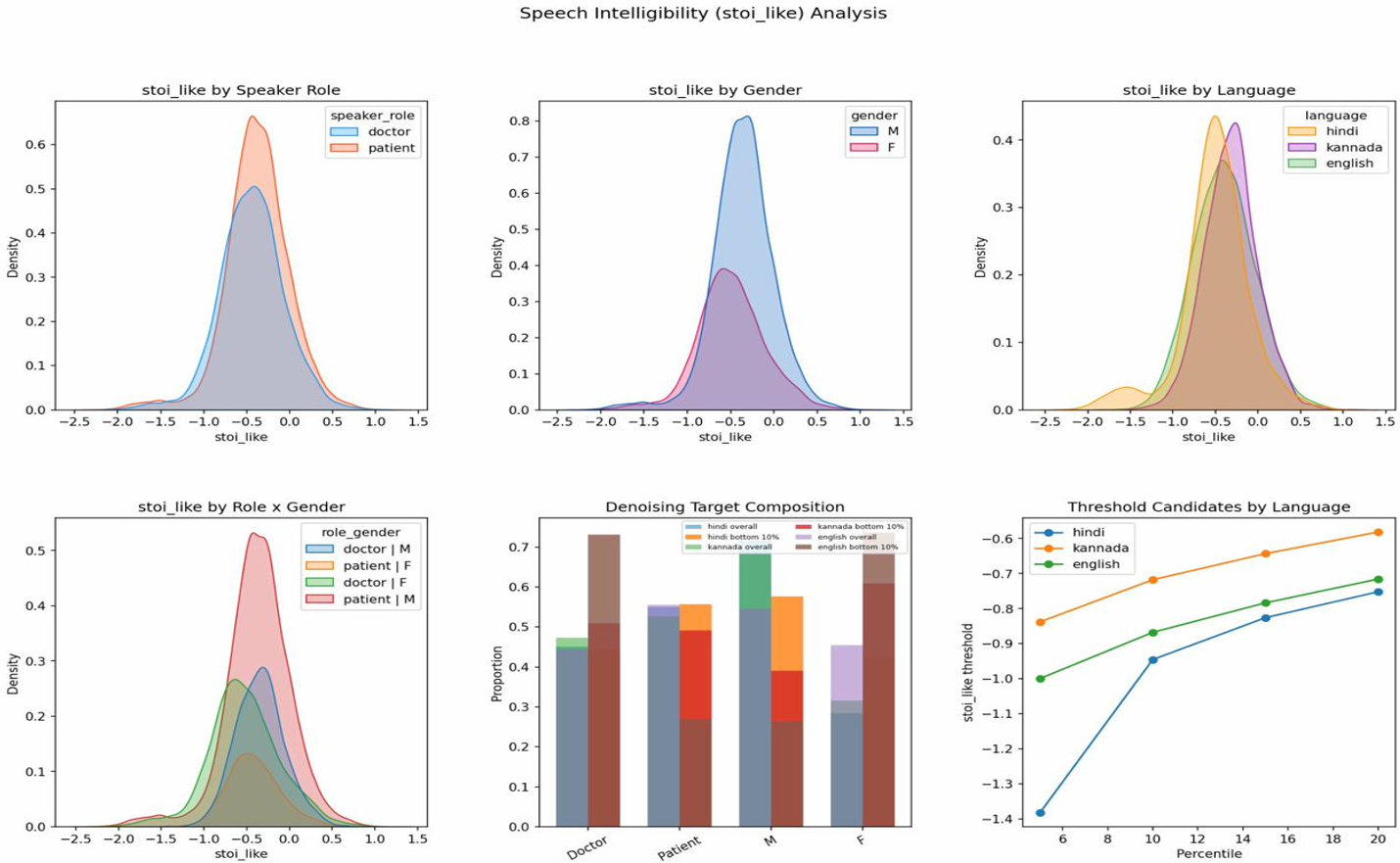}
  \caption{\footnotesize Speech intelligibility report of the speech data used in this study.}
  \label{fig:stoi}
\end{figure*}

In this section, we analyze the acoustic features of the dataset used in this stdy. We evaluate several key features, which indicates how good the voice quality of the data is. Performance evaluations indicate that males, patients, and Kannada speakers are structurally disadvantaged subgroups which are disproportionately affected by higher Word Error Rates ($\mathcal{WER}$). Importantly, the performance gap stems from acoustic differences, not data size. Although males represent 65\% of the dataset, their speech leads to worse WERs because of a naturally lower fundamental frequency ($r = 0.84$) and a lower voice quality, indicated by lower Harmonics-to-Noise Ratio (HNR) and higher amplitude instability (shimmer). Likewise, the speech of patients is a reflection of clinical realities such as psychomotor symptoms or emotional distress, which manifest acoustically as lower pitch and degraded voice quality. We depict the differences in pitch and voice quality of the speech data among gender and speaker role respectively from Figures ~\ref{fig:pitchbyrole} to ~\ref{fig:vcbygender}. In addition, the illustrate the speech intelligibility analysis in Figure ~\ref{fig:stoi}.

\onecolumn
\section{Annotation guidelines}
\label{apx:annotation_guidelines}
In this section, we detail the exact transcription instructions given to annotators with examples for each language.
\vspace{5mm}

\noindent\textbf{\textit{Transcription instructions to annotators}}:\\
Your task is to carefully listen to the provided audio file and create a \textbf{.txt} or \textbf{.srt} file.\\
  \textbf{About the recordings}:\\
  $\bullet$ These are audio-recorded clinical interviews between doctors and patients.\\
  $\bullet$ Accuracy of transcription is very important for clinical and research purposes.\\
  $\bullet$ Please follow the following guidance material to ensure accurate transcriptions.\\
  % \begin{itemize}
  %     \item These are audio-recorded clinical interviews between doctors and patients.
  %     \item Accuracy of transcription is very important for clinical and research purposes.
  %     \item Please follow the following guidance material to ensure accurate transcriptions.
  % \end{itemize}
Carefully read the instructions below on how to do the transcriptions and their examples for each of the three language, namely, English, Hindi, and Kannada (in order).\\
\textbf{Speaker diarisation}:\\
Clearly distinguish between the speakers. Use the following labels at the beginning of each new turn of speech.\\
$\bullet$ For English $\to$ [Doctor:] [Patient:]\\
$\bullet$ For Hindi $\to$ [\texthindi{डॉक्टर}:] [\texthindi{मरीज़}:]\\
$\bullet$ For Kannada $\to$ [\textkannada{ವೈದ್ಯ}:] [\textkannada{ರೋಗಿ}:]\\
% \begin{itemize}
%     \item For English $\to$ [Doctor:] [Patient:]
%     \item For Hindi $\to$ [\texthindi{डॉक्टर}:] [\texthindi{मरीज़}:]
%     \item For Kannada $\to$ [\textkannada{ವೈದ್ಯ}:] [\textkannada{ರೋಗಿ}:]
% \end{itemize}
\textbf{Use of punctuation}:\\
Allowed punctuations are full-stop, question mark, comma, ellipsis, em-dash and exclamation marks.\\
\begin{tabularx}{\textwidth}{p{5cm} p{11cm}}
\hline
\textbf{Punctuation} & \textbf{Guidance} \\ \hline
Full Stop (.) & Use for the end of a sentence. \\ \hline
Comma (,) & Use for short pauses and separating multiple items or phrases. \\ \hline
Ellipsis (...) & Use for long pauses. \\ \hline
Em-dash (--) & Use for interruptions or cut-offs. \newline
 English example, Doctor: So what brings you h-- \newline
 Hindi example, \texthindi{चिकित्सक: तुम यहाँ क्यों} -- \newline
 Kannada example, \textkannada{ಡಾಕ್ಟರ್: ನೀವು ಯಾಕೆ ಇಲ್ಲಿಗೆ} –- \\ \hline
Question Mark (?) & Use for direct questions. \\ \hline
Exclamation (!) & Use a single mark to indicate strong emotions. \newline
English example, It just feels so bad I can’t even tell you! \newline
Hindi example, \texthindi{मैं आपको बता नहीं सकता कि मैं कितना परेशान हूँ}! \newline
Kannada example, \textkannada{ನಂಗೆ ಎಷ್ಟು ಬೇಜಾರ್ ಆಗುತ್ತೆ ಅಂದ್ರೆ ಹೇಳೋಕ್ಕೆ ಆಗಲ್ಲ}! \\
\bottomrule
\end{tabularx}
\textbf{Strict verbatim transcription}:\\
Preserve all dysfluencies like pauses, filler words, stammers etc. Do not correct if there are grammatical errors in the speech itself. See table below. \\
\begin{tabular}{p{4.5cm} p{3.5cm} p{3.5cm} p{3.5cm}}
\toprule
\textbf{Guidance} & \textbf{English examples} & \textbf{Hindi examples} & \textbf{Kannada examples} \\
\midrule
Capture filler words and phrases 
& Um, Yeah, Hmm, Mmmm, etc 
& \texthindi{अच्छा, हाँ, हम्म, हा, आ, } etc 
& \textkannada{ಉಮ್, ಅಹ್, ಹಮ್ಮ್, ಹಾ, ಆ,} etc \\
\midrule
Include all incomplete phrases as they are said 
& I... I yesterday... no, the day before, I had gone there. 
& \texthindi{मैं...मैं कल...नहीं, परसों वहाँ गया था।} 
& \textkannada{ನಾನು... ನಾನು ನಿನ್ನೆ... ಇಲ್ಲ, ಮೊನ್ನೆ ಅಲ್ಲಿಗೆ ಹೋಗಿದ್ದೆ} \\
\midrule
Include stutters and stammers 
& I-I-I get scared. 
& \texthindi{मैं-मैं-मैं डर जाता हूं|} 
& \textkannada{ನ-ನ-ನನಗೆ ಭಯ ಆಗುತ್ತೆ} \\
\midrule
Use ellipsis (three dots) for long pauses or incomplete sentences 
& I don’t know… I feel very scared. 
& \texthindi{मुझे नहीं पता... मुझे बहुत डर लग रहा है|} 
& \textkannada{ನನಗೆ ಗೊತ್ತಿಲ್ಲ... ತುಂಬಾ ಭಯ ಆಗುತ್ತೆ} \\
\midrule
Transcribe as spoken, do not change informal to formal style (applies for languages like Kannada where spoken/colloquial and written styles are very different). Do not correct grammar or polish the output in any manner. 
& I am afraid $\to$ I am feeling scared
& \texthindi{मुझे डर है $\to$ मुझे डर लग रहा है} 
& \textkannada{ನಂಗೆ ಭಯ ಆಗ್ತಿದೆ $\to$ ನನಗೆ ಭಯ ಆಗುತ್ತಿದೆ} \\
\midrule
There can be frequent language mixing in these audios. You can expect English, Hindi, Telugu, Tamil and Malayalam. Write these words in their native script as of speech. 
& main samajh gaya, gottaytu
& \texthindi{मैं समझ गया, गोथायतू} 
& \textkannada{ಠೀಕ್ ಹೈ, ನಂಗೆ ಅರ್ಥ ಆಯಿತು} [theek hai, nange artha aytu] \\
\midrule
If there are numbers, type them out in words and not as numerals. 
& \textcolor{red}{Incorrect}: I took 10 tablets that day. \newline\textcolor{fg}{Correct}: I took ten tablets that day. 
& \textcolor{red}{Incorrect}: \texthindi{मैंने 10 गोलियाँ लीं / मैंने १० गोलियाँ लीं।}\newline
\textcolor{fg}{Correct}: \texthindi{मैंने दस गोलियाँ ले लीं }
& \textcolor{red}{Incorrect}: \textkannada{ನಾನು ಅವತ್ತು 10 ಮಾತ್ರೆಗಳು ತಗೊಂಡೆ / ನಾನು ಅವತ್ತು ೧೦ ಮಾತ್ರೆಗಳು ತಗೊಂಡೆ} \newline
\textcolor{fg}{Correct}: \textkannada{ನಾನು ಅವತ್ತು ಹತ್ತು ಮಾತ್ರೆಗಳು ತಗೊಂಡೆ} \\

% Example of spoken numerals 
% & My phone number is nine two zero seven
% & \texthindi{मेरा फ़ोन नंबर है नाइन टू जीरो सेवेन}
% & \textkannada{ನನ್ನ ಫೋನ್ ನಂಬರ್ ನೈನ್ ಟೂ ಜೀರೋ ಸೆವೆನ್} \\
\bottomrule
\end{tabular}
\\
% \newline\newline
% (contd.)

\noindent\textbf{Non-speech occurrences}:\\
Five relevant non-speech sounds must be noted. Use square brackets [$\dots$] for these annotations. For example, when the discussion is about depression, sounds of the speaker crying becomes important. See the indicators below. Do not use any other indicators.\\
\begin{tabular}{p{5cm} p{3.5cm} p{3.5cm} p{3.5cm}}
\toprule
\textbf{Type} & \textbf{English examples} & \textbf{Hindi examples} & \textbf{Kannada examples} \\
\midrule
Three Special tokens are allowed for emotional expressions. They must be in square brackets and are the only ones allowed. & [laughs], [cries], and [shouts] & [\texthindi{हंसने की आवाज़}], [\texthindi{रोने की आवाज़}] and [\texthindi{चिल्लाना}] & [\textkannada{ನಗುವಿನ ಸದ್ದು}], [\textkannada{ಅಳುವಿನ ಸದ್ದು}] and [\textkannada{ಕೂಗುತ್ತಾ}] \\
\hline
One special token is allowed for unclear speech & [unclear] & [\texthindi{अस्पष्ट}] &  [\textkannada{ಅಸ್ಪಷ್ಟ}] \\
\hline
One special token is allowed for other noises \textbf{[noise]}. Use this for all non-human background sounds like phone ringing or buzzing, ambulance siren, car honks, furniture scraping, other mechanical sounds. & [00:03:16] Patient: I had come then, but [noise] could not meet him. & [00:03:16] \texthindi{मरीज़: मैं तब आया था, लेकिन} [noise] \texthindi{मिला नहीं} & [00:03:16] \textkannada{ರೋಗಿ: ಆವಾಗ್ಲೇ ಬಂದಿದ್ದೆ, ಆದ್ರೆ} [noise] \textkannada{ಸಿಗ್ಲಿಲ್ಲ}.\\
\bottomrule
\end{tabular}

% Custom column: RaggedRight + pretolerance=10000 
% This stops LaTeX from wasting time trying to hyphenate Kannada/Hindi words.
\newcolumntype{P}[1]{>{\RaggedRight\arraybackslash\small\pretolerance=10000}p{#1}}

\noindent
\textbf{Timestamping and General Formatting: }\\
Use uniform font size and line spacing. The document must be in .txt or .srt format. The transcript must be timestamped in [HH:MM:SS] format (hours: minutes: seconds). See the example below:\\
\begin{tabular}{p{5cm} p{5cm} p{5cm}}
\toprule
\textbf{English examples} & \textbf{Hindi examples} & \textbf{Kannada examples} \\ \hline
% ROW 1
[00:00:05] Doctor: Please come in, have a seat, and tell me, how are you? \newline
[00:00:18] Patient: Hello, Doctor. I haven't been feeling well for the past one or two weeks and I feel sad all the time.\newline
[00:01:05] Patient: Yes, Doctor. I used to enjoy talking to my friends and gardening, but now I feel like just sitting alone all the time.
& 
[00:00:05] \texthindi{डॉक्टर: आइए, बैठिए और बताइए कि आप कैसे हैं}? \newline
[00:00:18] \texthindi{मरीज़: नमस्कार डॉक्टर साहब. पिछले एक-दो सप्ताह से मेरी तबीयत ठीक नहीं है और हमेशा उदास रहता हूं.} \newline
[00:01:05] \texthindi{मरीज़: हाँ डॉक्टर, पहले मुझे अपने दोस्तों से बात करना और बागवानी करना अच्छा लगता था, लेकिन अब हर समय अकेले बैठे रहने का मन करता है.}
& 
[00:00:05] \textkannada{ವೈದ್ಯರು: ಬನ್ನಿ, ಕೂತ್ಕೊಳ್ಳಿ. ಹೇಳಿ, ಈವಾಗ ಹೇಗಿದ್ದೀರ}? \newline
[00:00:18] \textkannada{ರೋಗಿ: ನಮಸ್ಕಾರ ಡಾಕ್ಟರ್. ಒಂದು ಎರಡು ವಾರದಿಂದ ಏನೋ ಸರಿ ಇಲ್ಲ. ಯಾವಾಗಲೂ ಬೇಜಾರು... ಒಂಥರಾ ಅನ್ಸುತ್ತೆ.} \newline
[00:01:05] \textkannada{ರೋಗಿ: ಹೌದು ಡಾಕ್ಟರ್. ಫ್ರೆಂಡ್ಸ್ ಜೊತೆ ಮಾತಾಡೋದು, ಗಿಡ ನೋಡಿಕೊಳ್ಳೋದು ಅಂದ್ರೆ ಇಷ್ಟ. ಆದ್ರೆ ಈಗ ಏನೂ ಮಾಡೋಕೆ ಇಷ್ಟ ಆಗಲ್ಲ. ಸುಮ್ಮನೆ ಕೂತಿರ್ತೀನಿ ಅಷ್ಟೇ.} \\
\bottomrule
\end{tabular}\\
A speaker’s entire turn must be kept in one line. Do not press ENTER (start a new line) in the middle of their speech even if it contains multiple sentences. Use line breaks between separate speakers. See example below.\\
\begin{tabular}{P{5cm} P{5cm} P{5cm}}
\toprule
\textbf{English examples} & \textbf{Hindi examples} & \textbf{Kannada examples} \\ \hline
\textcolor{fg}{\textbf{Correct}}\newline
Doctor: When was the last time you slept well?\newline
Patient: I can’t remember. It’s been many months. \newline
\textcolor{red}{\textbf{Incorrect}}\newline
Doctor: When was the last time you slept well?\newline
Patient: I can’t remember.\newline
It’s been many months.
& 
\textcolor{fg}{\textbf{Correct}}\newline
\texthindi{चिकित्सक: पिछली बार आपको अच्छी नींद कब आई थी}?\newline
\texthindi{मरीज़: मुझे याद नहीं. लगभग एक महीना हो गया.} \newline
\textcolor{red}{\textbf{Incorrect}}\newline
\texthindi{चिकित्सक: पिछली बार आपको कब अच्छी नींद आई थी}?\newline
\texthindi{मरीज़: मुझे याद नहीं.} \newline
\texthindi{लगभग एक महीना हो गया.}
& 
\textcolor{fg}{\textbf{Correct}}\newline
\textkannada{ವೈದ್ಯರು: ನೀವು ಕೊನೆಯದಾಗಿ ಯಾವಾಗ ಚೆನ್ನಾಗಿ ನಿದ್ದೆ ಮಾಡಿದ್ದೀರಿ?} \newline
\textkannada{ರೋಗಿ: ನನಗೆ ನೆನಪಿಲ್ಲ. ಸುಮಾರು ಒಂದು ತಿಂಗಳಾಯ್ತು.} \newline
\textcolor{red}{\textbf{Incorrect}}\newline
\textkannada{ವೈದ್ಯರು: ನೀವು ಕೊನೆಯದಾಗಿ ಯಾವಾಗ ಚೆನ್ನಾಗಿ ನಿದ್ದೆ ಮಾಡಿದ್ದೀರಿ?}\newline
\textkannada{ರೋಗಿ: ನನಗೆ ನೆನಪಿಲ್ಲ.}\newline
\textkannada{ಸುಮಾರು ಒಂದು ತಿಂಗಳಾಯ್ತು.} \\
\bottomrule
\end{tabular}